# Examining Redundancy in the Context of Safe Machine Learning

Hans Dermot Doran
ZHAW / Institute of Embedded Systems
Winterthur, Switzerland
donn@zhaw.ch

Monika Reif
ZHAW / Institute of Applied Mathematics and Physics
Winterthur, Switzerland
reif@zhaw.ch

*Abstract*—**This paper describes a set of experiments with neural network classifiers on the MNIST database of digits. The purpose is to investigate naïve implementations of redundant architectures as a first step towards safe and dependable machine learning. We report on a set of measurements using the MNIST database which ultimately serve to underline the expected difficulties in using NN classifiers in safe and dependable systems.**

## I. Introduction

### A. Motivation

The motivation for this body of work is, in the context of machine learning (ML) in safe and dependable systems, to examine integration of Neural Network (NN) classifiers in safe and dependable systems.

Machine learning and safe and dependable systems are not mutually exclusive and we have several initiatives where we seek to combine the two. The first is a high-speed functionally safe machine monitoring system where the switch-off, i.e. entry into the safe state must occur within 10 microseconds. In such systems it is advantageous to anticipate system failure and pre-emptively initiate a shut-down to reduce start-up and repair costs. In a system of 12'000 sensors this can almost only be achieved using ML. In this case only availability suffers if a faulty prediction is made (False Negatives.)

A second case is compensating for lossy and redundant sensors in process control by anticipating sensor values. In this application ML can predict what the sensors output value could be. In this case the use of ML is also non critical - despite a critical process – because the output value can be bounded and indeed ignored if an a-priori ruleset is compromised.

The problem with neural networks, specifically those utilised as classifiers, is that they are architected to produce a best-effort result. In fact most neural networks do not feature a default "don't know" output and therefore there are no True Negatives [1]. The effect is that the available outcomes are a True Positive or a False Positive. We seek to use NN classifiers in order to distinguish artefacts that are difficult to describe using mathematical models and instead are transferred into the probabilistic domain to achieve a result that is ultimately based on some abstraction of a comparison. The objective is to achieve as many matches as possible from uncertain and a priori unknown inputs. In functionally safe systems, the objective is to avoid dangerous situations which, in classification, typically arise from False Positives.

The novelty we present is a preliminary investigation of the application of standard high-integrity techniques and architectures onto neural network classifiers. Typically redundancy and diversity techniques are used to gain a definitive answer from a number of sensors, calculations and/or algorithms diverse or not, where one or more sensors or operations may fail. If we allow that NN classifiers are ultimately sensors we are faced with a number of possibilities, two of which we wish to consider here namely same and diverse redundancy. Redundancy is often used to ensure that a calculation has been performed correctly and to achieve this certainty we would consider using two same NN classifiers with the same training data. Reproducibility is a minimum expectation and can be trivially demonstrated.

Diversity is often used to avoid common-mode and other systematic failures and what we wish to investigate is the naive viewpoint that two diverse classifiers can be used to confirm the correct functioning of the classifier. Redundancy is often associated with voting, which is failure masking. Diversity in sensors where each sensor outputs a systematically different but comparable output, rarely use any mathematical voting function more complex than an arithmetic or logical function. Indeed, more complex functions generally fall into the category of sensor fusion whose objective is another so a research question is, how can the outputs of classifiers be voted on.

We approach these challenges in a practical pragmatic way by undertaking a set of experiments, because of which we dispense with a section on previous work, and structure the rest of the paper accordingly. In the Section II we describe the experiments and in Section III we draw conclusions.

## II. EXPERIMENTS

In this section we describe the performance and results of a set of experiments investigating standard redundancy techniques. As an experimental platform we use the FINN library from Xilinx [2] which implements the two versions of NN we use, a Fully Connected Network (LRC) and the Convolutional Neural Network (CVR) as binary neural networks.

The FINN suite supplies a LRC pre-trained with the 60'000 28*28 pixel images from the MNIST database [3] and we use these as the basis for further experiments. For the CVR machine we need to both adapt the images to 32*32, which we do by extending the border, and re-train the machines. We re-train the networks on a GPU cluster provided by a sister institute [4] and in order to do so must adapt the setup for Singularity rather than the Docker container supplied by the FINN repository.

In order to determine the correctness of the classification the test data were compared with the results of the classification on a filename basis.

### A. Experiment 1 – Baseline Data

We re-train the two networks successively with 60'000 and 1'000 images networks on a GPU cluster and present the test set of 10'000 to both.

Table 1 below shows the True Positive/False Positive percentages as well as the classification times measured on the PYNQ platform.

Table 1: Baseline of correctly and incorrectly classified images by number of training images as well as the classification times per variant. W1 standard for 1 weight bit, A1 for one activation bit.

|  | LFC W1 A1 | LFC W1 A2 | CNV W1 A1 | CNV W1 A2 | CNV W2 A2 |
|---|---|---|---|---|---|
| 60'000 Images |  |  |  |  |  |
| True Positives | 98.4% | 98.49% | 99.58% | 99.51% | 99.61% |
| False Positives | 1.6% | 1.51% | 0.42% | 0.49% | 0.39% |
|  |  |  |  |  |  |
| 1'000 Images |  |  |  |  |  |
| True Positives | - | - | 95.41% | 97.06% | 97.26% |
| False Positives | - | - | 4.59% | 2.94% | 2.74% |
|  |  |  |  |  |  |
| Classification Time | 8 us | 8 us | 328 us | 328 us | 1161 us |

The results for the LFC possess less meaning than one might hope. The output is a binary vector, that is a 10 bit vector representing the digits (0 ... 9) where one index has logical value "1" and the rest logical value "0." This decision is made in the interface between the HW classifier and the SW output module on the basis of first highest value. That is if two digits are identified with the same probability, the earlier digit is chosen. This issue is also possible, albeit less likely given the bit-size of the weighting produced by the CNV. In order to assess the likely impact, we examined the occurrence of the same weight for each image from the test data set. As the results show for the three CNV configurations trained with 60'000 images, there are no occurrences of two or more weights being equal (Table 2.)

Table 2: Number of equal weights output by CNV configurations trained with different numbers of training data

|  | CNV W1A1 | | CNV W1A2 | | CNV W2A2 | |
|---|---|---|---|---|---|---|
| Trained part | 1'000 | 60'000 | 1'000 | 60'000 | 1'000 | 60'000 |
| Equal Weightings | 8 | 0 | 6 | 0 | 3 | 0 |

As previously mentioned, the fundamentally serious issue is that there is no possibility to categorise Negatives, a not recognised image. A False Positive is, in terms of functional safety, a dangerous failure.

We test the static reproducibility of the networks. That is we present the same image to the trained network 10'000 times and our expectation that the same result – for the CNV network that all the weightings are equal every time - is achieved, is validated. Whilst this test does not verify that the NN does not possess internal states it does help show that these states does not necessarily affect the output.

We also test the reproducibility of the networks in that we change the order of presentation of training data. The training machine receives labelled data in a single file of ubyte format and presents this data to the training implementation which calculates the convolution matrixes on an epoch by epoch basis. Our expectation is that the order of data presentation to the training algorithm does not affect the output, an expectation validated by this experiment.

### B. Experiment 2 Logical Voting

As previously stated, voting serves to mask errors and here we investigate the output of parallel. The general 2-out-of-2 (2oo2) architecture is shown in Figure 1 below. As previously explained there are two forms. In the first form, Neural Network 1 and Neural Network 2 are identical. For each input they should output an identical value and the voter's task is to confirm that this is the case. The redundancy serves to verify that the classification was carried out correctly. Having shown that the reproducibility of the neural networks under test is given, we eschew an experiment with two similar neural networks, as we do not expect to observe anything but random errors, specifically Single Event Upsets (SEUs) rooted in environmental action on the HW.

We therefore focus on diversity in implementation where Neural Network 1 and Neural Network 2 (Figure 1) differ in some fashion. This configuration reduces the scope for systematic errors, often at the expense of voter simplicity.

The act of voting in this configuration expands the possible outputs. Outputs are either a Positive or an Undecided, the latter being that the two networks disagree.

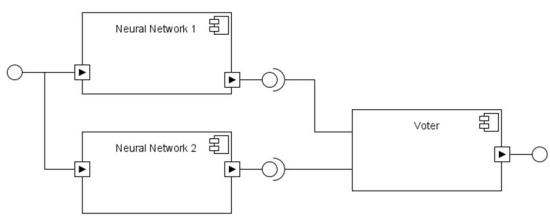

Figure 1: General pattern of a physical redundant system

In the case of Positives, post-hoc investigation can then reveal whether the Positive was either correctly or falsely identified. There are therefore three mutually exclusive output data sets. The set of True Positives, the set of False Positives and the Set of Undecideds. These three sets are listed in Table 3, Table 4 and Table 5 below. In this experiment we might (naively) expect that the value for True Positives defaults to the lowest value of the two neural networks. This would indicate that the other neural network recognises a super-set of the first. This is unfortunately not the case, each NN recognises a subset of the other as a True Positive. The issue now becomes either a philosophical question with regards to the meaning of ground truth or an exercise in fully probabilistic design. Neither bode well for functionally safe systems.

Table 3: True Positive rates for logic voting for the two NNs with differing weights (W) and Activation (A) bits. The fields with grey background are values gained from Experiment 1. The outlined box is referred to in Table 6 further below.

|  | LFC W1A1 | LFC W1A2 | CNV W1A1 | CNV W1A2 | CNV W2A2 |
|---|---|---|---|---|---|
| LFC W1A1 | 98,40 % | 98,15 % | 98,23 % | 98,22 % | 98,24 % |
| LFC W1A2 | 98,15 % | 98,49 % | 98,31 % | 98,29 % | 98,34 % |
| CNV W1A1 | 98,23 % | 98,31 % | 99,58 % | 99,33 % | 99,46 % |
| CNV W1A2 | 98,22 % | 98,29 % | 99,33 % | 99,51 % | 99,38 % |
| CNV W2A2 | 98,24 % | 98,34 % | 99,46 % | 99,38 % | 99,61 % |

Table 4: False Positive rates for logic voting for the two NNs with differing weights (W) and Activation (A) bits. The fields with grey background are values gained from Experiment 1.

|  | LFC W1A1 | LFC W1A2 | CNV W1A1 | CNV W1A2 | CNV W2A2 |
|---|---|---|---|---|---|
| LFC W1A1 | 1,60 % | 1,12 % | 0,19 % | 0,25 % | 0,22 % |
| LFC W1A2 | 1,12 % | 1,51 % | 0,17 % | 0,23 % | 0,22 % |
| CNV W1A1 | 0,19 % | 0,17 % | 0,42 % | 0,21 % | 0,24 % |
| CNV W1A2 | 0,25 % | 0,23 % | 0,21 % | 0,49 % | 0,25 % |
| CNV W2A2 | 0,22 % | 0,22 % | 0,24 % | 0,25 % | 0,39 % |

Table 5: Undecided rates for logic voting for the two NNs with differing weights (W) and Activation (A) bits.

|  | LFC W1A1 | LFC W1A2 | CNV W1A1 | CNV W1A2 | CNV W2A2 |
|---|---|---|---|---|---|
| LFC W1A1 | - | 0,73 % | 1,58 % | 1,53 % | 1,54 % |
| LFC W1A2 | 0,73 % | - | 1,52 % | 1,48 % | 1,44 % |
| CNV W1A1 | 1,58 % | 1,52 % | - | 0,46% | 0,30% |
| CNV W1A2 | 1,53 % | 1,48 % | 0,46% | - | 0,37% |
| CNV W2A2 | 1,54 % | 1,44 % | 0,30% | 0,37% | - |

### C. Experiment 3 Arithmetic Voting

As previously pointed out the standard output of the LFC is a binary vector so redundancy can only converge to a True/False output. The CNV can output a weight and two weights can be arithmetically connected together to produce a stronger predicate. We investigate two forms of redundancy here firstly the case of two CNV networks with different activation and weight bits and two CNV networks with the same number of activation and weight bits but different training sets.

#### 1) Experiment 3a - Diversity in CNV Weightings/Activations

By simple addition of two weighted outputs of two diverse CVNs we may (naively) expect an improvement on the diversity with logical voting. This form of voting will however only grant us two possible results, True and False Positives. The table below Table 6 therefore only notes the True Positives and is the result of addition of the output weights. From the results and comparing with the heavy bordered box from Table 3 we can see that there is a definite improvement in True Positives. Given that we do not really understand the behaviour of Experiment 2 because we do not understand the relationship between True and False positives, we consider this result to be of secondary importance.

Table 6: True Positive rates for arithmetic voting for the two NNs with differing weights (W) and Activation (A) bits. The fields with grey background are values gained from Experiment 1. The values can be compared with the outlined box in Table 2. We note, for instance, an improvement of 0.3% in True Positives between Logical and Arithmetic voting for the CNVA1W1 and CNV W1A2 (diverse) configuration.

|  | CNV W1A1 | CNV W1A2 | CNV W2A2 |
|---|---|---|---|
| CNV W1A1 | 99,58 % | 99,63 % | 99,63 % |
| CNV W1A2 | 99,63 % | 99,51 % | 99,66 % |
| CNV W2A2 | 99,63 % | 99,66 % | 99,61 % |

### 2) Experiment 3b – Diversity of training

In this experiment we train the same NN networks with a diverse training set. Specifically we split the training set in two sets of 30'000 images each, train two same CVN networks, present the networks with the test set and note the results (Table 7.) We learn that the True Positives come very close to that of the networks trained with all 60'000 training images. As the table shows, the True Positive rates of the single networks are weaker. We feel that this result indicates a path forward using unmodified NN architectures

Table 7: Individual True Positive rates for diversely trained (30'000 images each) same networks also compared with arithmetically combined; voted True Positive rates compared with True Positive Rates from Experiment 1.

|  | CNV W1A1 | | CNV W1A2 | | CNV W2A2 | |
|---|---|---|---|---|---|---|
| Trained part | 1/2 | 2/2 | 1/2 | 2/2 | 1/2 | 2/2 |
| True Positive Rate | 99,42% | 99,56% | 99,41% | 99,44% | 99,45% | 99,3% |
| True Positives | 99.57% | | 99.50% | | 99.57% | |
| True Positive Rate Experiment 1 | 99.58% | | 99.51% | | 99.61% | |

### D. Applying Thresholds

By measuring the distance between a likely correct classification and the next-best-fit, we achieve some sort of confidence factor and if we use that confidence factor as a threshold we can also map the output of a NN to True Positives and Uncertains. In an attempt to evaluate a useful threshold size we tracked the threshold size against True/False Positives and Uncertains, the results are shown in Figure 2. The diagram shows a marginal reduction in False Positives at the cost of an increase in Uncertains. The one False Positive-crossing threshold is different for every CNV parametrisation.

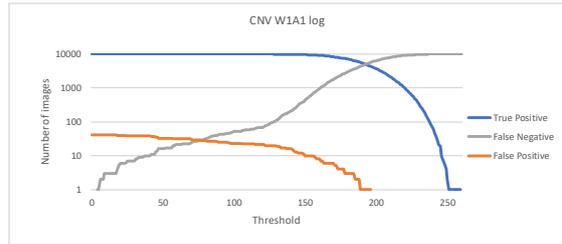

Figure 2: Log plot of True Positives, False Positives and Uncertains against thresholds for the CNV parametrized with one activation and one weight bit.

### E. Experiment 4 - Temporal Redundancy

Neural network classifiers are generally stateless, that is classifications have no memory. Temporal redundancy, running the same algorithm twice on the same input data, is another technique to ensure that the algorithm is performed correctly. Diversity in implementations are also common, again used to avoid systematic errors in the implementation.

### F. Experiment 5 - Data Redundancy

In a similar but different vein, a diverse input, that is two versions of the same image for instance possibly with changed X-Y position, rotated or different lighting, may possibly be used to generate a True Positive or Uncertain out of an otherwise potential False Positive. We conduct two general experiments on rotation. The first examines the effect of variations on True Positives, the second on False Positives. shows the rotation of a True Positive image of the digit 8 which is broadly representative of the remaining digits in that rotations plus minus 10 degrees generally, but not exclusively, make no difference to the prospects of recognition by a trained (in this case CNV W1A1) network.

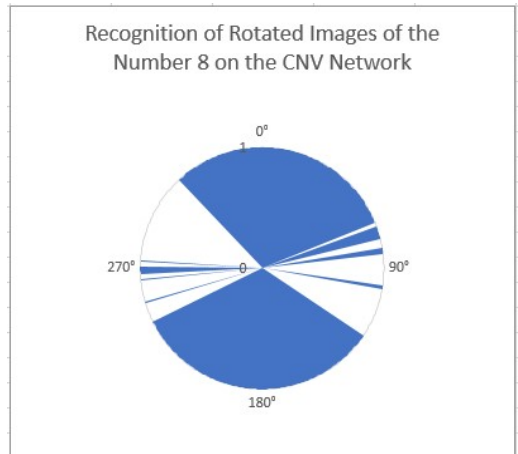

Figure 3: CNV recognition of the digit 8 rotated through 360°. Blue indicates recognized digit.

What is arguably more interesting is whether, for instance, rotating those digits that are not recognised by a trained network will improve recognition chances. From Experiment 1, there are 11 images of the number 8 that are not recognised. We take those 11 images and rotate them in one degree steps through 360°, passing them individually through the CNV network. Figure 4 shows us recognition rates of images that are rotated by the degrees that are noted on the X-axis. After rotation to some degree, all digits can be recognised some of the time, there are no False Positives detected any more. What the figure tells us, for instance, is that 10 images of the digit 8 could be recognised when rotated by ca. 340°. It does not tell us the distribution of possible rotations for a specific digit.

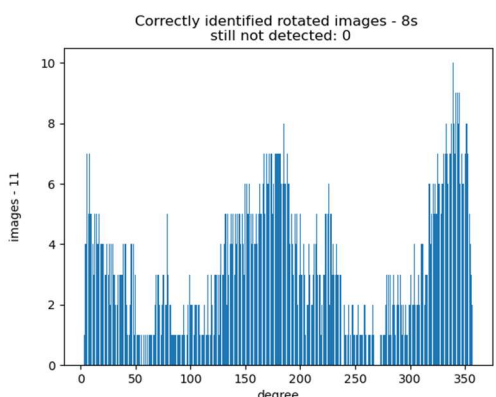

Figure 4: Recognition of previously unrecognised images of the digit 8 by degree of rotation

Figure 5 below shows us the recognition for a single exemplary image of the digit 8 is rotated through 360°.

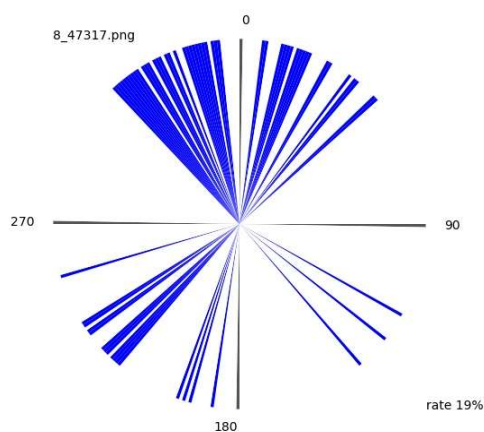

Figure 5: Recognition of a previously unrecognised digit 8 through rotation. Blue denotes recognised digit.

### III. DISCUSSION

We must again re-iterated that we must be careful not to confuse concepts. Utilising redundancy is a technique for dependability leading to high integrity operation. Redundancy is in its usual sense, applied on units of known operation models, for instance temperature sensors whose operation is well understood or processors where the code has been vetted. The issue with neural network classifiers is that the model enacted on the neural network is generally not well understood, if at all and highly susceptible to bias. While we can expect a redundant execution, under error-free conditions, of a neural network classifier to produce the same result, the reaction of the redundant execution with respect to a broad set of input-data is indeterminate and may produce dangerous operational situations. A large part of this problem is caused by an inability, with current classifier architectures, to put precise bounds on the training data that allow prediction of results.

This paper examines the question of redundancy as a technique to reduce the potential for dangerous operations, with limited success. The value of this paper lies in the discussion of some safety and dependability issues which may help trigger a more theoretical discussion on the use of NN classifiers in functionally safe systems.

### ACKNOWLEDGMENTS

Thanks are due to Colin Dreher and Alexander Lau for their work in setting up and performing a large part of the experiments in the framework of a student thesis [5] as well as Thorvin Stasiak for some additional experimental work.